# Evolutionary Multi-Objective Optimization Framework for Mining Association Rules

## Shaik Tanveer Ul Huq[1,2] and Vadlamani Ravi[1*]

[1]Center of Excellence in Analytics,
Institute for Development and Research in Banking Technology,
Castle Hills Road #1, Masab Tank, Hyderabad – 500 057, INDIA.
[2] School of Computer & Information Sciences,
University of Hyderabad, Hyderabad – 500 046, INDIA
shaiktanveerulhuq@gmail.com ; rav_padma@yahoo.com;

*Abstract*— **In this paper, two multi-objective optimization frameworks in two variants (i.e., NSGA-III-ARM-V1, NSGA-III-ARM-V2; and MOEAD-ARM-V1, MOEAD-ARM-V2) are proposed to find association rules from transactional datasets. The first framework uses Non-dominated sorting genetic algorithm III (NSGA-III) and the second uses Decomposition based multi-objective evolutionary algorithm (MOEA/D) to find the association rules which are diverse, non-redundant and non-dominated (having high objective function values). In both these frameworks, there is no need to specify minimum *support* and minimum *confidence*. In the first variant, *support*, *confidence*, and *lift* are considered as objective functions while in second, *confidence*, *lift*, and *interestingness* are considered as objective functions. These frameworks are tested on seven different kinds of datasets including two real-life bank datasets. Our study suggests that NSGA-III-ARM framework works better than MOEAD-ARM framework in both the variants across majority of the datasets.**

## I. INTRODUCTION

As technology is continuously upgrading rapidly, interactions among people and the technology are also growing. Consequently, a humongous amount of data representing the interactions or transactions among entities is generated day by day. To analyze this massive data, several data mining techniques were proposed. Association rule mining, one of the data mining tasks introduced by Agrawal [1], is used to extract knowledge from the transactional data. The examples of transactional data are bank transactions, customer transactions at Supermarket, and so on. Some of the applications of association rule mining include Market Basket Analysis, Medical Diagnosis, Protein sequence, Census data, and fraud detection in credit card business and cyber fraud detection etc [36]. Association rule mining algorithms extract association rules from transactional data. Applications of Association rule mining includes auditing file system permissions [29], detecting and predicting software defects [30] [31], finding influential users in social media [32], Outlier detection [33] and in recommendation systems [34] etc. Association rules help us represent what items/products are dependent on each other. An association rule is represented by X → Y, where X is a set of products, and Y is another set of products such that X ∩ Y =ϕ, i.e., products present in set Y should not be present in set X and vice versa. Here this association rule X → Y tells that when any customer consumes/buys the products present in set X, there is a high chance that he/she consumes the product Y also. Since 1993 plenty of research has been done in this area.

* Corresponding Author; Phone:+914023294310; FAX: +914023535157

There are mainly three types of association rule mining they are Categorical or Binary Association rule mining, Quantitative association rule mining, and fuzzy association rule mining [13] [14],. In transactional databases, the rows represent transactions and columns represents products. In categorical association rule mining, the transactional dataset contains only two values i.e., 0 or 1, '0' means the corresponding product is not present in the transaction and '1' means the corresponding product is present in the transaction. In Quantitative association rule mining, the transactional dataset can contain any continuous value. For example, if n is present anywhere in the dataset, then it means the corresponding product is used n number of times in the corresponding transaction. But as this numerical association mining deal with sharp boundaries between consecutive intervals, it can't represent smooth changes in between consecutive intervals which can be easily taken care by fuzzy association rule mining. This paper focuses on binary association rule mining.

In 1993, Agarwal [1] proposed an algorithm called Apriori algorithm, which uses metrics such as *support* and *confidence* (explained in the later section) to extract the association rules from the categorical transactional dataset. This Apriori algorithm works in two phases. In the first phase it generates set of frequently occurring products or items in the transactions using *support* metric, and in the next phase, it generates the reliable rules from set obtained in the first phase using *confidence* metric. In 2000, Han [2] proposed the FP-growth algorithm, which uses the divide and conquer method to generate association rules. The above two methods need minimum *support* and minimum *confidence* values to be specified upfront. Later, many optimization frameworks have been proposed to extract association rules. In 2013, Sarath and Ravi [3] had proposed binary particle swarm optimization framework which does not need the minimum *support* and minimum *confidence* values to extract the association rules. In 2014, Pradeep and Ravi [4] considered this association rule mining problem as a multi-objective optimization problem and proposed three frameworks i.e. MO-BPSO, MO-BFFOO-TA and MO-BPSO-TA . These algorithms  generated top 10 association rules and does not require minimum *support* and minimum *confidence* by considering the fitness function as the product of *support (or coverage)*, *confidence*, *Interestingness*, *Comprehensibility*, *Lift*, *Leverage*, and *conviction* metrics.

Recently in the literature, many evolutionary algorithms have been used to solve this association rule mining problem. References [15], [16], [17] and [18] used Genetic algorithm and [20],[21] and [22] used particle swarm optimization to solve association rule mining problem.  Reference [19] used a hybrid model comprising of both genetic algorithm and particle swarm optimization algorithm to mine association rules. Then, Bat evolutionary algorithm was applied to to generate association rules ([23, [24], [25]). Later, Wolf search algorithm to mine association rules from transactional data [26]. Then, references [27] and [28] used modified binary cuckoo search algorithm and Niche-Aided Gene Expression Programming respectively to generate association rules from the transactional datasets.

In this paper, we proposed two multi-objective optimization frameworks, i.e. NSGA-III-ARM and MOEA/D-ARM in two variants to obtain non-redundant, non-dominated, diversified rules.

This rest of the paper is organized as follows: Section II describes the Motivation; Section III presents the Contribution, Section IV describes the basic definitions needed, Section V presents the proposed methodology, Section VI describes the datasets used here, Section VII presents Experimental Analysis, Results and Discussion, and finally, Section VIII concludes the paper.

## II. Motivation

There are several metrics such as *support (*also called *coverage)*, *confidence*, *lift*, *interestingness*, *Comprehensibility*, *Leverage* and *conviction* etc., which describe the Association Rules obtained by using any known framework. In literature, most of the frameworks identified the association rule mining problem as a multi-objective optimization problem but solved it in a single objective or bi-objective environment. When solved using single objective optimization algorithms, some devised the fitness function involving some of the metrics mentioned above, mostly, *support* and *confidence*. However, this approach is fraught with disadvantages. For instance, in such approaches, we may not obtain (i) diverse rules (ii) rules which are optimal with respect to other important measures. In this paper, we devised a framework which helps the decision maker by yielding a set of diverse and non-redundant rules. The same is the case with works that proposed bi-objective framework for the problem. Therefore, there is a pressing need to come out with a framework, where all important measures are considered as objective functions and the resulting problem is solved with multi-objective optimization algorithms.

## III. Contributions

In this paper, two new multi-objective optimization frameworks, i.e. NSGA-III-ARM and MOEA/D-ARM are proposed in two variants to extract diverse, non-redundant association rules from categorical transactional data.

- In the first variant, the objective functions considered are *support*, *confidence*, and *lift*. Optimizing *Support* and *Confidence* values alone may not remove misleading strong associations. Hence lift is also included as the third objective which tells us about the correlations between antecedent and consequent sets in an association rule.

- In the second variant, the objective functions considered are *confidence*, *lift*, and *interestingness*. Traditional Association rule miners like Apriori algorithm usually generates rules based on frequent item-sets. But it may fail if we want the rules based on rare item-sets. Addition of *interestingness* measure as an objective function gives the rules based on rare item-sets. Thus, considering *confidence*, lift and *interestingness* as objective functions which need to be optimized gives the strong rare association rules.

- A new way of generating solutions has been implemented when the dataset is too sparse.

IV. BASIC DEFINITIONS

***Metrics for calculating the strength of an association rule:***

a. ***Support***: *Support* is defined as the percentage or fraction of transactions in the database that contain items present in both antecedent as well as the consequent sets.

$$Support(A \rightarrow B) = \frac{\text{Frequency}(AUB)}{\text{N}}$$

b. ***Confidence***: *Confidence* indicates how reliable or relevant a given rule is. *Confidence* is defined as the probability of occurring the rule's consequent under the condition that the transactions also contain the antecedent.

$$Confidence(A \rightarrow B) = \frac{\text{Frequency}(AUB)}{\text{Frequency(A)}}$$

c. ***Lift***: The *lift* value is a measure of the importance of a rule (originally called interest). Lift tells us about the correlation between antecedent and consequent in the association rule. If the lift value is less than 1, then it means that there is a negative correlation between antecedent and consequent. If the lift value is 1, then it means that there is no correlation between antecedent and consequent. If the lift of an association rule is greater than 1, then it means that there is a positive correlation between antecedent and consequent.

$$Lift(A \rightarrow B) = \frac{Confidence\,(A \rightarrow B)}{Support\,(B)} = \frac{Support\,(A) \times\ Support\,(B)}{Support\,(B)}$$

d. ***Interestingness***: A rule is said to be interesting when the individual *support* count values are greater than the collective *support* (A→B) values. If the item's present in the association rule are rare item's(items which are participated in very less transactions in the given dataset) then it's *interestingness* value will be high and if the item's present in the association are frequent items (items which are participated in more number of transactions in the given dataset) then its' interesting value will be low.

$$Interestingness = \frac{Support\,(AUB)}{Support\,(A)} \times \frac{Support\,(AUB)}{Support\,(B)} \times (1 - \frac{Support\,(AUB)}{Support\,(D)})$$

## V. Proposed Methodology

### *Problem Formulation:*

We formulate the association rule mining problem as a multi-objective optimization problem in two variants. In the first variant, we considered *Support*, *Confidence*, and *Lift* as objective functions, and in the second variant, we considered *Confidence*, *Lift*, and *Interestingness* as objective functions. The basic block diagram of the flow of our proposed frameworks is as below. In view of the definitions provided in section IV, lift is a very important measure in determining the strength of the association between antecedent and consequent of a rule in that support and confidence alone do not contain that information. On the other hand, support and interestingness are poles apart in that a rule with high support indicates its frequency whereas a rule with high interestingness will have low support which implies that they are rare rules. In some applications, rare rules are also important. Therefore, it is futile to include both support and interestingness as objective functions in the model. By including interestingness as an objective function, we want to mine those rare rules also.

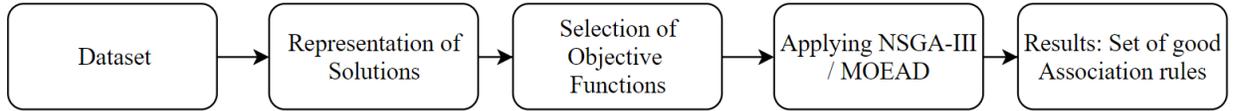

Fig 1 Schematic of proposed frameworks

### *Rule Representation or Rule Encoding:*

Michigan approach [5] has been used in this paper to represent a solution or an association rule or chromosome. In this representation, if the number of products in the dataset is N, then the length of the solution represented will be 2N, where each item is represented by two consecutive bits. Each bit can take the values either 0 or 1. If an item's two consecutive bits value is '11' then it means that the item is present in the antecedent, else if the value is '10' then it means that the item is present in the consequent, else if the values are '00' or '01', then it means that the item is absent in the rule. For programming simplicity, we considered the solution as array of size N, where N is the number of products in any given transactional dataset. In this array each cell can have three values 0, 1 and 2. If the $i^{th}$ cell value is 0 then it means that the $i^{th}$ product is in the antecedent in the association rule, else if the $i^{th}$ cell value is 1, then it means that the $i^{th}$ product is in the consequent part of association rule and if the $i^{th}$ cell value is 2, then it means that $i^{th}$ product is not in the rule. An example representation of an association rule is depicted in Fig. [1]. Here there are five products i.e., $I_1$, $I_2$, $I_3$, $I_4$ and $I_5$, and the corresponding values are 0, 2, 0, 0, 1. It means that $I_1$, $I_3$, $I_4$ items are present in the antecedent part and $I_5$ is present in the consequent part of the association rule. $I_2$ is not present in the rule at all as its corresponding value is 2.

| I1 | I2 | I3 | I4 | I5 |
|----|----|----|----|----|
| 0  | 2  | 0  | 0  | 1  |

Fig 2 Example of Representation of Association Rule

***NSGA-III Algorithm:***

In the first framework, NSGA-III-ARM, we used the Non-dominated Sorting Genetic Algorithm III proposed by Deb et al. [6]. It is a multi and many objective evolutionary optimization algorithm designed to optimize problems having 3 to 15 objectives. Generally, when the number of objectives increases, the number of non-dominated solutions also increases. This algorithm yields a set of solutions that are converged and have diversity among them. This property helps the decision maker by providing a number of different options (or solutions). In this algorithm, the reference-based framework is used to select a set of solutions from a large number of non-dominated solutions to maintain diversity in the selected solutions.

The input for this algorithm is a population of randomly initialized solutions and reference or aspiration points. Here reference points are the points which are uniformly taken from the plane in objective space with the intercept on all axes. Initially, we randomly generate a population containing N solutions using a particular representation. After that, we compute the objective function values of these solutions. Then, we perform crossover and mutation to get the child population with other N solutions. We then again compute the objective function values of the solutions in the child population. Thus, we have 2N solutions, and we need to select the best N solutions out of 2N. We apply non-dominated sorting on 2N solutions which gives us the Pareto fronts $Pf_1$, $Pf_2$…etc. Out of these Pareto fronts, we select first 'l' Pareto fronts such that the sum of several solutions present in those first 'l' Pareto fronts, i.e. $Pf_1$, $Pf_2$, … $Pf_l$ is greater than or equal to population size, i.e. N. If this sum is exactly equal to N, then the solutions present in Pareto fronts $Pf_1$, $Pf_2$, … $Pf_l$ are considered as the parent population for the next iteration. However, if that sum is greater than N, then we select the solutions present in Pareto fronts $Pf_1$, $Pf_2$… $Pf_{l-1}$ and we needed to select k = N - | Pf$_1$ U Pf$_2$ U … Pf$_{l-1}$| solutions from the last Pareto front Pfl for the next iteration. For selecting k solutions from the last Pareto front, we do three things. First one is normalizing the objective functions of all solutions in Pf$_1$, Pf$_2$… Pf$_l$. Then, we plot all these solutions in the objective function space. Second, we draw reference lines, i.e., lines from the origin to the reference points which are initially given as input and each solution present in Pf$_1$, Pf$_2$… Pf$_l$ Pareto fronts are associated with the reference line that is nearest to it. Third, we calculate the niche count of each reference line. Niche count is the number of solutions assigned to the reference line. After that, we take k solutions present in the last Pareto front Pf$_l$ one at a time whose niche count is less to get the diverse solutions. This process is repeated until the pre-specified number of iterations are completed.

### *MOEAD Algorithm:*

MOEA/D [7] is another multi-objective optimization algorithm which decomposes a multi-objective problem into several scalar objective optimization problem and optimizes those scalar objectives. Several decomposition methods are reported in literature. Some of the well-known approaches are Weighted sum approach, Tchebycheff Approach [8], and Penalty based boundary intersection approach [9]. In the experiments conducted, we have used Penalty based boundary intersection approach as a decomposition method over Tchebycheff approach as it gives more uniformly distributed solutions when the objectives are more than two when the objectives are more than two[7].

The basic MOEA/D algorithm works as follows. The inputs needed for the MOEA/D algorithm are multi-objective optimization problem, N, λ, T, and a stopping criterion, where N is the number of scalar subproblems, λ = (λ$_1$, λ$_2$, … λ$_N$) is a set of N weight vectors. T is the number of weight vectors in the neighborhood of each weight vector.

In the initialization phase, we generate N solutions randomly or by any other means, calculate objective function values for all N solutions. We associate each of those N solutions to any one of the weight vectors.. For each weight vector, calculate its T nearest weight vectors using Euclidean distance. Then we initialize the ideal solution z as (z$_1$,z$_2$,…z$_m$) where z$_i$ is the best value of objective 'i' found so far and m is the number of objective functions that need to be optimized.

After initialization, repeat the four steps, i.e., Reproduction, Update of z, Update of Neighbouring Solutions steps for every solution till the stopping criterion is satisfied. In the reproduction step of any solution a, we randomly select two solutions b, c whose associated weight vectors are present in the T nearest weight vectors of a weight vector associated with solution a. After selecting those two solutions, we perform genetic operations such as crossover and mutation on both solutions b, c to produce new solution d. After that we update ideal solution z as follows: for minimization problem if $z_i < F_i(d)$ then set $z_i = F_i(d)$ where $F_i(d)$ it $i^{th}$ objective function value of solution d. In the update of Neighbouring solutions step, we select each solution 'x' associated with the T nearest weight vectors of weight vector associated with solution a, and set x = d if g(d| $\lambda_x$,z) <= g(x| $\lambda_x$,z), where 'g' is the scalar optimization function defined by the decomposition method and $\lambda_x$ is the weight vector associated with solution x.

After the stopping criterion is satisfied, the non-dominated solutions obtained from the obtained solution set are considered as the output of MOEA/D algorithm.

When we do crossover or mutation operations, there is a chance of producing invalid or infeasible solutions. Hence, we added a repair function in both the frameworks which checks for the invalid rules generated by them and replace them with a randomly generated rule.

***Customizations Effected in NSGA-III and MOEAD frameworks***

***Changes effected  in Initialization:***

Most of the time, binary transactional datasets are very sparse, i.e. the overall number of products are more when compared to the average number of products in each transaction. For those type of datasets, if we randomly generate the N solutions, then there will be a high probability that the generated solutions are invalid, i.e. *support* becomes zero. Picking better solutions in the initialization phase greatly helps the evolutionary algorithms. Thus we have randomly chosen N transactions from the dataset and considered them as initial solutions or rules by randomly picking any one product present in that transaction and placed in the consequent part of the rule, and the remaining products present in that transaction is placed in the antecedent part of the rule. For the Clickstream, XYZ bank, PQR bank datasets (description given in the next section), we used this type of initialization, and for remaining datasets, we randomly generated the N solutions.

***Adding repair function after the generation of child solutions:***

While conducting experiments, we kept two constraints, i.e. no redundant or duplicate rules should be present in the population, and each solution or association rule should contain only one item or product in the consequent. Antecedent can contain one or more products that are not present in the consequent. For that purpose, we have added a repair function after the child population is generated. When we apply crossover and mutation operations on parent population, there is a chance that the produced children contain zero or two items present in the consequent part, which does not satisfy the above-mentioned constraints. For that purpose, if there are two items present in the consequent part, randomly one item will be transferred to antecedent, and if no items are present in the consequent part then one randomly picked an item from the antecedent part is transferred to consequent part.

***Evaluation Functions and Measure of Convergence and Diversity.***

When true Pareto front is not known, an approximate true Pareto front can be calculated by running an evolutionary algorithm with large population size and a large number of iterations [10]. Thus, to know the approximate true Pareto front of our association rule mining problem, we ran our NSGA-III-ARM framework for both variants by considering high population size and a high number of generations, i.e. 500 and 500 respectively. After approximating the true Pareto front, we computed Inverted Generational Distance (IGD) [11] and Hypervolume (HV) [12] values which are respectively used to measure the diversity and convergence of solutions obtained by any multi-objective frameworks. Lower the IGD value of a framework with a particular configuration; the nearer is the Pareto front obtained by that framework for that particular configuration to the true Pareto front. Higher the HV value of a framework with a particular configuration, the more diverse are the solutions obtained by that framework for that particular configuration.

IGD (Inverted Generational Distance) is computed as follows:

$$IGD(A, Z_{eff}) = \frac{1}{|Z_{eff}|} \sum_{i=1}^{|Z_{eff}|} \min_{j=1}^{|A|} d(z_i, a_j)$$

Here, $d(z_i, a_j) = \left\| z_i - a_j \right\|_2$, where A is the set of solutions obtained by the algorithm, $Z_{eff}$ is the set of points present in Pareto optimal surface. $a_j$ is a solution present in set A. $z_i$ is a solution in the Pareto optimal surface which is near to $a_j$.

The Hypervolume of set X is the volume of space formed by non-dominated points present in set X with any reference point. Here the reference point is the "worst possible" point or solution (any point that is dominated by all the points present in solution set X) in the objective space.

## VI. DATASET DESCRIPTION

We analyzed seven datasets in this paper. They are Books, Food, Grocery, two real life Bank datasets (PQR bank and XYZ bank), Bakery and Clickstream.. The Books dataset is taken from [www.solver.com/xlminer-data-mining](www.solver.com/xlminer-data-mining) containing 11 types of books and 2,000 consumer records. Food dataset is taken from [www.ibm.com/software/analytics/spss,](www.ibm.com/software/analytics/spss) which contains 11 kinds of foods and 1,000 number of transactions. Grocery dataset is taken from [http://www.sas.com/technologies/analytics/datamining/miner](http://www.sas.com/technologies/analytics/datamining/miner) containing 20 grocery items and 1001 transaction records. Bakery dataset is taken from [https://wiki.csc.calpoly.edu/datasets/wiki/ExtendedBakery,](https://wiki.csc.calpoly.edu/datasets/wiki/ExtendedBakery) which consists of 40 different kinds of pastries and ten kinds of coffee drinks and with 1,000 transactions. The Clickstream dataset is taken from [http://archive.ics.uci.edu/ml/datasets/Anonymous+Microsoft+Web+Data](http://archive.ics.uci.edu/ml/datasets/Anonymous+Microsoft+Web+Data) containing the information of 37,711 undisclosed, randomly picked users who visited [www.microsoft.com](www.microsoft.com) website and surfed through 294 types of website hyperlinks present on that website.

## EXPERIMENTAL SETTINGS

### A. System Configuration

The experiments were executed on a system having Intel Xeon(R) CPU E5-2640 v4 2.4 GHz, with eight cores and 32 GB RAM in Ubuntu 16.04 environment. The code for both variants was developed in language Python 3.6 and used libraries such as pymoo, pymop, numpy, pandas etc. in it.

### B. Parameter Settings

The number of generations or iterations for both frameworks and their variants is fixed at 200. The crossover probabilities chosen are 0.8, 0.85, and 0.9. The mutation probabilities considered are 0.1, 0.15, 0.2. The number of runs for each parameter combination is 30. For NSGA-III framework variants, the population size considered as 50 and the number of reference points we have taken is 91. For MOEA/D framework variants the number of reference points considered as 45 and the population size is the same as the number of reference points. The number of Neighbors is taken as 20, and the decomposition method used is Penalty based boundary intersection method.

## VII. RESULTS AND DISCUSSION

To know which framework performed well among the two, i.e. NSGA-III-ARM, MOEA/D-ARM in two variants each, we calculated the ratio of average of Hypervolume to Inverted Generational Distance, i.e., HV/IGD values obtained for 30 runs among all above-mentioned parameter combinations for both the frameworks for all the datasets. The highest average HV/IGD values for all datasets for all variants is shown in Table 1 and Table 2. The top 10 frequent rules obtained for 30 runs for the best parameter combinations mentioned in Table 1 and Table 2 are presented in Tables 3-30. The time complexity of the NSGA-III algorithm is $O(N^2M)$ and that of MOEA/D is $O(NMT)$ where N is the population size, M is the number of objective functions and T is the number of solutions considered as the neighbour solutions in MOEA/D algorithm. Despite having higher time complexity, NSGA-III outperformed MOEA/D in almost all cases.

**BakeryMod:**

The highest average HV/IGD value obtained for NSGA-III-ARM-V1 is more than that of MOEAD-ARM-V1, as shown in Table 1 for the Bakery Dataset. The frequencies of the rules obtained for 30 runs for NSGA-III-ARM-V1 (shown in Table 4) are from 7 to 3, which is better when compared to MOEAD-ARM-V1 i.e., 1(shown in Table 3). From this, we can clearly say that MOEAD-ARM-V1 is not giving the same rules consistently when compared to NSGA-III-ARM-V1 for the given parameter combinations.

The highest average HV/IGD value obtained for NSGA-III-ARM-V2 is more than that of MOEAD-ARM-V2, as shown in Table 2 for the Bakery Dataset. The frequencies of the rules obtained for 30 runs for NSGA-III-ARM-V2 (shown in Table 6) are from 6 to 2, which is better when compared to MOEAD-ARM-V2 i.e., 6 to 2(shown in Table 5). From this, we can clearly say that MOEAD-ARM-V2 is not giving the same rules consistently when compared to NSGA-III-ARM-V2 for the given parameter combinations.

**ClickStream:**

The highest average HV/IGD value obtained for NSGA-III-ARM-V1 is less than that of MOEAD-ARM-V1, as shown in Table 1 for the Bakery Dataset. The frequencies of the rules obtained for 30 runs for NSGA-III-ARM-V1 (shown in Table 8) are from 24 to 3 which is same and similar when compared to MOEAD-ARM-V1 i.e., 24 to 3(shown in Table 7). From this, we can clearly say that MOEAD-ARM-V1 is not giving the same rules consistently when compared to NSGA-III-ARM-V1 for the given parameter combinations.

The highest average HV/IGD value obtained for NSGA-III-ARM-V2 is more than that of MOEAD-ARM-V2, as shown in Table 2 for the Bakery Dataset. The frequencies of the rules obtained for 30 runs for NSGA-III-ARM-V2 (shown in Table 10) are from 21 to 3 which is same and similar when compared to MOEAD-ARM-V2 i.e., 21 to 3 (shown in Table 9). From this, we can clearly say that MOEAD-ARM-V2 is not giving the same rules consistently when compared to NSGA-III-ARM-V2 for the given parameter combinations.

**Grocery:**

The highest average HV/IGD value obtained for NSGA-III-ARM-V1 is more than that of MOEAD-ARM-V1, as shown in Table 1 for the Bakery Dataset. The frequencies of the rules obtained for 30 runs for NSGA-III-ARM-V1 (shown in Table 16) are from 30 to 30, which is better when compared to MOEAD-ARM-V1 i.e., 3 to 1 (shown in Table 15). From this, we can clearly say that MOEAD-ARM-V1 is not giving the same rules consistently when compared to NSGA-III-ARM-V1 for the given parameter combinations. Six new rules are obtained by NSGA-III-ARM-V1 and Ten new rules are generated by MOEAD-ARM-V1 framework when compared with the rules generated by the BPSO framework [3] for Books dataset because of the addition of the third objective i.e., *Lift* in addition to *support* and *confidence*. The new rules have been indicated in bold letters in Table 15 and Table 16 respectively.

The highest average HV/IGD value obtained for NSGA-III-ARM-V2 is more than that of MOEAD-ARM-V2, as shown in Table 2 for the Bakery Dataset. The frequencies of the rules obtained for 30 runs for NSGA-III-ARM-V2 (shown in Table 18) are from 29 to 1, which is better when compared to MOEAD-ARM-V2 i.e., 1 (shown in Table 17). From this, we can clearly say that MOEAD-ARM-V2 is not giving the same rules consistently when compared to NSGA-III-ARM-V2 for the given parameter combinations.

**Food:**

The highest average HV/IGD value obtained for NSGA-III-ARM-V1 is more than that of MOEAD-ARM-V1, as shown in Table 1 for the Bakery Dataset. The frequencies of the rules obtained for 30 runs for NSGA-III-ARM-V1 (shown in Table 20) are from 30 to 30, which is better when compared to MOEAD-ARM-V1 i.e., 5 to 1(shown in Table 19). From this, we can clearly say that MOEAD-ARM-V1 is not giving the same rules consistently when compared to NSGA-III-ARM-V1 for the given parameter combinations. Seven new rules are obtained by NSGA-III-ARM-V1 and Ten new rules are generated by MOEAD-ARM-V1 framework when compared with the rules generated by the BPSO framework [3] for Books dataset because of the addition of the third objective i.e., *Lift* in addition to *support* and *confidence*. The new rules have been indicated in bold letters in Table 19 and Table 20 respectively.

The highest average HV/IGD value obtained for NSGA-III-ARM-V2 is more than that of MOEAD-ARM-V2, as shown in Table 2 for the Bakery Dataset. The frequencies of the rules obtained for 30 runs for NSGA-III-ARM-V2 (shown in Table 22) are from 30 to 30, which is better when compared to MOEAD-ARM-V2 i.e., 5 to 1(shown in Table 21). From this, we can clearly say that MOEAD-ARM-V2 is not giving the same rules consistently when compared to NSGA-III-ARM-V2 for the given parameter combinations.

**Books:**

The highest average HV/IGD value obtained for NSGA-III-ARM-V1 is more than that of MOEAD-ARM-V1, as shown in Table 1 for the Bakery Dataset. The frequencies of the rules obtained for 30 runs for NSGA-III-ARM-V1 (shown in Table 24) are from 30 to 30, which is better when compared to MOEAD-ARM-V1 i.e., 4 to 1(shown in Table 23). From this, we can clearly say that MOEAD-ARM-V1 is not giving the same rules consistently when compared to NSGA-III-ARM-V1 for the given parameter combinations. Ten new rules are obtained by NSGA-III-ARM-V1 and Ten new rules are generated by MOEAD-ARM-V1 framework when compared with the rules generated by the BPSO framework [3] for Books dataset because of the addition of the third objective i.e., *Lift* in addition to *support* and *confidence*. The new rules have been indicated in bold letters in Table 23 and Table 24 respectively. They are obviously more practical and useful than the ones reported in Sarath and Ravi [3].

The highest average HV/IGD value obtained for NSGA-III-ARM-V2 is more than that of MOEAD-ARM-V2, as shown in Table 2 for the Bakery Dataset. The frequencies of the rules obtained for 30 runs for NSGA-III-ARM-V2 (shown in Table 26) are from 30 to 30, which is better when compared to MOEAD-ARM-V2 i.e., 5 to 2(shown in Table 25). From this, we can clearly say that MOEAD-ARM-V2 is not giving the same rules consistently when compared to NSGA-III-ARM-V2 for the given parameter combinations.

**XYZ bank:**

The highest average HV/IGD value obtained for NSGA-III-ARM-V1 is less than that of MOEAD-ARM-V1, as shown in Table 1 for the Bakery Dataset. The frequencies of the rules obtained for 30 runs for NSGA-III-ARM-V1 (shown in Table 28) are from 22 to 6, which is better when compared to MOEAD-ARM-V1 i.e., 15 to 4(shown in Table 27). From this, we can clearly say that MOEAD-ARM-V1 is not giving the same rules consistently when compared to NSGA-III-ARM-V1 for the given parameter combinations. Eight new rules are obtained by NSGA-III-ARM-V1 and Eight new rules are generated by MOEAD-ARM-V1 framework when compared with the rules generated by the BPSO framework [3] for Books dataset because of the addition of the third objective i.e., *Lift* in addition to *support* and *confidence*. The new rules have been indicated in bold letters in Table 27 and Table 28 respectively.

The highest average HV/IGD value obtained for NSGA-III-ARM-V2 is more than that of MOEAD-ARM-V2, as shown in Table 2 for the Bakery Dataset. The frequencies of the rules obtained for 30 runs for NSGA-III-ARM-V2 (shown in Table 30) are from 9 to 1 which is same and similar when compared to MOEAD-ARM-V2 i.e., 9 to 2(shown in Table 29). From this, we can clearly say that MOEAD-ARM-V2 is giving the same rules consistently when compared to NSGA-III-ARM-V2 for the given parameter combinations.

**PQR Bank:**

The highest average HV/IGD value obtained for NSGA-III-ARM-V1 is more than that of MOEAD-ARM-V1, as shown in Table 1 for the Bakery Dataset. The frequencies of the rules obtained for 30 runs for NSGA-III-ARM-V1 (shown in Table 12) are from 30 to 6, which is better when compared to MOEAD-ARM-V1 i.e., 6 to 1(shown in Table 11). From this, we can clearly say that MOEAD-ARM-V1 is not giving the same rules consistently when compared to NSGA-III-ARM-V1 for the given parameter combinations.

The highest average HV/IGD value obtained for NSGA-III-ARM-V2 is more than that of MOEAD-ARM-V2, as shown in Table 2 for the Bakery Dataset. The frequencies of the rules obtained for 30 runs for NSGA-III-ARM-V2 (shown in Table 14) are from 15 to 4, which is better when compared to MOEAD-ARM-V2 i.e., 8 to 2 (shown in Table 13). From this, we can clearly say that MOEAD-ARM-V2 is not giving the same rules consistently when compared to NSGA-III-ARM-V2 for the given parameter combinations.

Thus, for both variants, NSGA-III framework obtained the highest HV/IGD value when compared to the MOEA/D framework except two datasets in the second variant (i.e., XYZ bank and Clickstream datasets). Thus, we can say from results that NSGA-III framework outperformed MOEAD framework in both the variants.

In both frameworks, addition of one extra objective function compared to the extant literature, indeed did the trick in obtaining better rules, with high practical significance. This is the significant outcome of the study.

## VIII. Conclusions

Two multi-objective optimization frameworks with two variants have been proposed (i.e., NSGA-III-ARM-V1, NSGA-III-ARM-V2, MOEAD-ARM-V1, MOEAD-ARM-V2) to extract association rules from transactional datasets. Results show that NSGA-III-ARM-V1 outperformed MOEAD-ARM-V1 in terms of the ratio of Hypervolume and Inverted Generational distance in all seven datasets. Further, we observed that the NSGA-III-ARM-V2 outperformed MOEAD-ARM-V2 in terms of the ratio of Hypervolume and Inverted Generational distance in five out of seven datasets. According to the criterion of the frequency of occurrence of the same rule when the algorithms were repeated for 30 runs, NSGA-III-ARM performed well by generating the rules consistently even ran at different times when compared to MOEAD-ARM in both variants. In addition to these, both the frameworks yielded us the non-redundant and diversified association rules, and there is no need to pre-scpecify the minimum *support* and *confidence* to obtain association rules.

Table 1 Average hv/igd value for variant1

| Problem | Framework | prob_cross | prob_mut | hv/igd |
|---|---|---|---|---|
| BakeryMod | MOEAD-ARM-V1 | 0.9 | 0.2 | 57.1066 |
| | **NSGA-III-ARM-V1** | **0.8** | **0.1** | **1157.1124** |
| ClickStream | MOEAD-ARM-V1 | 0.8 | 0.2 | 185.7812 |
| | **NSGA-III-ARM-V1** | **0.8** | **0.2** | **219.8293** |
| Grocery (Grocery) | MOEAD-ARM-V1 | 0.9 | 0.1 | 121.6963 |
| | **NSGA-III-ARM-V1** | **0.9** | **0.1** | **374089.5641** |
| Food(Food) | MOEAD-ARM-V1 | 0.8 | 0.1 | 151.6004 |
| | **NSGA-III-ARM-V1** | **0.8** | **0.2** | **Inf** |
| Books(Books) | MOEAD-ARM-V1 | 0.8 | 0.2 | 54.6981 |
| | **NSGA-III-ARM-V1** | **0.9** | **0.2** | **508229.6805** |
| XYZ bank | MOEAD-ARM-V1 | 0.9 | 0.2 | 107.8981 |
| | **NSGA-III-ARM-V1** | **0.9** | **0.1** | **175.1945** |
| PQR Bank | MOEAD-ARM-V1 | 0.8 | 0.2 | 77.1508 |
| | **NSGA-III-ARM-V1** | **0.9** | **0.1** | **2133.9244** |

Table 2 Average hv/igd values for variant 2

| problem | Framework | prob_cross | prob_mut | hv/igd |
|---|---|---|---|---|
| BakeryMod | MOEAD-ARM-V2 | 0.9 | 0.2 | 35.5071 |
| | **NSGA-III-ARM-V2** | **0.8** | **0.1** | **470.8232** |
| ClickStream | **MOEAD-ARM-V2** | **0.8** | **0.2** | **5231.3125** |
| | NSGA-III-ARM-V2 | 0.8 | 0.2 | 93.3400 |
| Grocery | MOEAD-ARM-V2 | 0.9 | 0.2 | 26.6479 |
| | **NSGA-III-ARM-V2** | **0.8** | **0.2** | **inf** |
| Food | MOEAD-ARM-V2 | 0.9 | 0.2 | 30.1281 |
| | **NSGA-III-ARM-V2** | **0.8** | **0.2** | **inf** |
| Books | MOEAD-ARM-V2 | 0.9 | 0.1 | 186.8063 |
| | **NSGA-III-ARM-V2** | **0.8** | **0.2** | **inf** |
| XYZ bank | **MOEAD-ARM-V2** | **0.8** | **0.2** | **1689.7271** |
| | NSGA-III-ARM-V2 | 0.9 | 0.1 | 1.9824 |
| PQR Bank | MOEAD-ARM-V2 | 0.9 | 0.2 | 242.4438 |
| | **NSGA-III-ARM-V2** | **0.9** | **0.1** | **337.2512** |

Table 3 MOEAD-ARM-V1-BakeryMod

| Freq | Antecedent | Consequent | *Support* | *Confidence* | *Lift* |
|---|---|---|---|---|---|
| 1 | Chocolate_T1,Casino,Blueberry_T1,Apricot_T3,Chocolate_T6 | Blackberry | 0.0010 | 1.0000 | 13.7000 |
| 1 | Napoleon,Lemon_T3,Raspberry_T2 | Pecan | 0.0010 | 1.0000 | 25.0000 |
| 1 | Blackberry,Tuile,Chocolate_T5,Raspberry_T2,Chocolate_T6 | Chocolate_T2 | 0.0010 | 1.0000 | 29.4000 |
| 1 | Coffee,Vanilla_T1,Almond_T3,Hot | Walnut | 0.0010 | 1.0000 | 16.4000 |
| 1 | Vanilla_T1,Walnut,Apricot_T2,Chocolate_T6 | Pecan | 0.0010 | 1.0000 | 25.0000 |
| 1 | Almond_T1,Lemon_T2,Apricot_T3 | Vanilla_T1 | 0.0010 | 1.0000 | 27.0000 |
| 1 | Ganache,Lemon_T3,Chocolate_T5,Orange,Bottled | Vanilla_T1 | 0.0010 | 1.0000 | 27.0000 |
| 1 | Lemon_T3,Chocolate_T5,Orange,Bottled | Vanilla_T1 | 0.0010 | 1.0000 | 27.0000 |
| 1 | Napoleon,Ganache,Lemon_T3,Raspberry_T2 | Pecan | 0.0010 | 1.0000 | 25.0000 |

Table 4 NSGA-III-ARM-V1-BakeryMod

| Frequency | Antecedent | Consequent | *Support* | *Confidence* | *Lift* |
|---|---|---|---|---|---|
| 7 | Napoleon | Strawberry | 0.049 | 0.5444 | 5.9829 |
| 6 | Truffle | Gongolais | 0.058 | 0.5631 | 5.2140 |
| 6 | Chocolate_T1 | Chocolate_T6 | 0.047 | 0.5595 | 6.5826 |
| 5 | Casino | Chocolate_T1 | 0.040 | 0.5556 | 6.6138 |
| 4 | Casino,Chocolate_T6 | Chocolate_T1 | 0.038 | 0.9744 | 11.5995 |
| 3 | Coffee,Almond_T3,Hot | Apple_T1 | 0.024 | 1.0000 | 14.7059 |
| 3 | Raspberry_T1,Green | Lemon_T3 | 0.019 | 0.9500 | 14.3939 |
| 3 | Chocolate_T1,Chocolate_T6 | Casino | 0.038 | 0.8085 | 11.2293 |
| 3 | Cherry_T1 | Apricot_T3 | 0.046 | 0.5476 | 7.3016 |
| 3 | Opera,Cherry_T1 | Apricot_T3 | 0.038 | 0.9268 | 12.3577 |

Table 5 MOEAD-ARM-V2-BakeryMod

| Freq | Antecedent | Consequent | Confidence | Lift | Interestingness |
|---|---|---|---|---|---|
| 1 | Chocolate_T1,Casino,Blueberry_T1,Apricot_T3,Chocolate_T6 | Blackberry | 1.0000 | 13.7000 | 0.0137 |
| 1 | Chocolate_T1,Blackberry,Blueberry_T1,Gongolais,Apricot_T3 | Casino | 1.0000 | 13.9000 | 0.0139 |
| 1 | Almond_T1,Lemon_T2,Apricot_T3 | Vanilla_T1 | 1.0000 | 27.0000 | 0.0270 |
| 1 | Ganache,Lemon_T3,Chocolate_T5,Orange,Bottled | Vanilla_T1 | 1.0000 | 27.0000 | 0.0270 |
| 1 | Lemon_T3,Chocolate_T5,Orange,Bottled | Vanilla_T1 | 1.0000 | 27.0000 | 0.0270 |
| 1 | Napoleon,Ganache,Lemon_T3,Raspberry_T2 | Pecan | 1.0000 | 25.0000 | 0.0250 |
| 1 | Napoleon,Lemon_T3,Blueberry_T2,Raspberry_T2 | Pecan | 1.0000 | 25.0000 | 0.0250 |
| 1 | Napoleon,Lemon_T3,Raspberry_T2 | Pecan | 1.0000 | 25.0000 | 0.0250 |
| 1 | Apple_T1,Ganache,Orange | Green | 1.0000 | 16.1000 | 0.0161 |
| 1 | Apple_T1,Almond_T3,Hot,Single | Blueberry_T2 | 1.0000 | 18.2000 | 0.0182 |

Table 6 NSGA-III-ARM-V2-BakeryMod

| Frequency | Antecedent | Consequent | Confidence | Lift | Interestingness |
|---|---|---|---|---|---|
| 6 | Casino,Chocolate_T6 | Chocolate_T1 | 0.9744 | 11.5995 | 0.4408 |
| 4 | Apple_T2,Apple_T3,Cherry_T2 | Apple_T4 | 1.0000 | 11.9048 | 0.3690 |
| 4 | Vanilla_T1,Cheese | Ganache | 0.7500 | 17.0455 | 0.0511 |
| 4 | Apple_T3,Apple_T4 | Apple_T2 | 0.9524 | 12.0555 | 0.4822 |
| 3 | Chocolate_T1,Chocolate_T6 | Casino | 0.8085 | 11.2293 | 0.4267 |
| 3 | Apple_T2,Apple_T4 | Apple_T3 | 0.9756 | 10.7210 | 0.4288 |
| 2 | Vanilla_T1,Cheese,Orange | Ganache | 1.0000 | 22.7273 | 0.0455 |
| 2 | Almond_T1,Apricot_T3 | Vanilla_T1 | 0.6667 | 18.0180 | 0.0360 |
| 2 | Vanilla_T1,Almond_T1,Gongolais,Apricot_T3 | Chocolate_T2 | 1.0000 | 29.4000 | 0.0294 |
| 2 | Coffee,Apple_T1,Hot | Almond_T3 | 1.0000 | 15.3846 | 0.3692 |

Table 7 MOEAD-ARM-V1-ClickStream

| Frequency | Antecedent | Consequent | *Support* | *Confidence* | *Lift* |
|---|---|---|---|---|---|
| 24 | L1034 | L1008 | 0.1608 | 0.5606 | 1.6923 |
| 8 | L1025 | L1026 | 0.0353 | 0.5440 | 5.5268 |
| 6 | L1008 | L1034 | 0.1608 | 0.4854 | 1.6923 |
| 5 | L1003 | L1001 | 0.0552 | 0.6085 | 4.4719 |
| 5 | L1026 | L1025 | 0.0353 | 0.3587 | 5.5268 |
| 4 | L1008,L1009,L1035 | L1018 | 0.0204 | 0.9074 | 5.5686 |
| 4 | L1041 | L1026 | 0.0282 | 0.6140 | 6.2374 |
| 4 | L1017,L1034 | L1008 | 0.0317 | 0.6701 | 2.0229 |
| 3 | L1009,L1017 | L1037 | 0.0168 | 0.4312 | 12.1582 |
| 3 | L1026 | L1041 | 0.0282 | 0.2860 | 6.2374 |

Table 8 NSGA-III-ARM-V1-ClickStream

| Frequency | Antecedent | Consequent | *Support* | *Confidence* | *Lift* |
|---|---|---|---|---|---|
| 24 | L1034 | L1008 | 0.1608 | 0.5606 | 1.6923 |
| 8 | L1025 | L1026 | 0.0353 | 0.5440 | 5.5268 |
| 6 | L1008 | L1034 | 0.1608 | 0.4854 | 1.6923 |
| 6 | L1026 | L1025 | 0.0353 | 0.3587 | 5.5268 |
| 5 | L1009 | L1037 | 0.0324 | 0.2293 | 6.4648 |
| 5 | L1003 | L1001 | 0.0552 | 0.6085 | 4.4719 |
| 5 | L1008,L1009,L1035 | L1018 | 0.0204 | 0.9074 | 5.5686 |
| 4 | L1017,L1034 | L1008 | 0.0317 | 0.6701 | 2.0229 |
| 4 | L1041 | L1026 | 0.0282 | 0.6140 | 6.2374 |
| 3 | L1026 | L1041 | 0.0282 | 0.2860 | 6.2374 |

Table 9 MOEAD-ARM-V2-ClickStream

| Frequency | Antecedent | Consequent | *Confidence* | *Lift* | *Interestingness* |
|---|---|---|---|---|---|
| 21 | L1034 | L1008 | 0.5606 | 1.6923 | 0.2721 |
| 6 | L1025 | L1026 | 0.5440 | 5.5268 | 0.1951 |
| 5 | L1008 | L1034 | 0.4854 | 1.6923 | 0.2721 |
| 4 | L1008,L1009,L1035 | L1018 | 0.9074 | 5.5686 | 0.1134 |
| 3 | L1026 | L1025 | 0.3587 | 5.5268 | 0.1951 |
| 3 | L1009,L1017 | L1037 | 0.4312 | 12.1582 | 0.2037 |
| 3 | L1017,L1034 | L1008 | 0.6701 | 2.0229 | 0.0642 |
| 3 | L1041 | L1026 | 0.6140 | 6.2374 | 0.1756 |
| 3 | L1009,L1018 | L1035 | 0.6386 | 11.6632 | 0.3352 |
| 3 | L1003 | L1001 | 0.6085 | 4.4719 | 0.2469 |

Table 10 NSGA-III-ARM-V2-ClickStream

| Frequency | Antecedent | Consequent | *Confidence* | *Lift* | *Interestingness* |
|---|---|---|---|---|---|
| 21 | L1034 | L1008 | 0.5606 | 1.6923 | 0.2721 |
| 6 | L1025 | L1026 | 0.5440 | 5.5268 | 0.1951 |
| 5 | L1008,L1009,L1035 | L1018 | 0.9074 | 5.5686 | 0.1134 |
| 5 | L1008 | L1034 | 0.4854 | 1.6923 | 0.2721 |
| 4 | L1026 | L1025 | 0.3587 | 5.5268 | 0.1951 |
| 3 | L1008,L1018,L1035 | L1009 | 0.8263 | 5.8404 | 0.1189 |
| 3 | L1009,L1017 | L1037 | 0.4312 | 12.1582 | 0.2037 |
| 3 | L1003 | L1001 | 0.6085 | 4.4719 | 0.2469 |
| 3 | L1041 | L1026 | 0.6140 | 6.2374 | 0.1756 |
| 3 | L1009,L1018 | L1035 | 0.6386 | 11.6632 | 0.3352 |

Table 11 MOEAD-ARM-V1-PQR Bank

| Frequency | Antecedent | Consequent | *Support* | *Confidence* | *Lift* |
|---|---|---|---|---|---|
| 6 | P66 | P71 | 0.0310 | 0.9146 | 26.6787 |
| 5 | P41,P71 | P66 | 0.0008 | 1.0000 | 29.5000 |
| 4 | P21 | P20 | 0.0087 | 0.7780 | 29.9000 |
| 4 | P6 | P18 | 0.0058 | 0.7000 | 39.4000 |
| 2 | P56 | P31 | 0.0029 | 0.5000 | 41.7000 |
| 2 | P20,P66,P71 | P47 | 0.0008 | 0.4000 | 40.4000 |
| 2 | P20,P66 | P47 | 0.0012 | 0.3750 | 37.8000 |
| 2 | P1,P2,P4,P17 | P49 | 0.0004 | 0.3330 | 29.9000 |
| 1 | P41 | P48 | 0.0017 | 0.1480 | 44.8000 |
| 1 | P20,P47,P71 | P66 | 0.0008 | 1.0000 | 29.5000 |

Table 12 NSGA-III-ARM-V1-PQR Bank

| Frequency | Antecedent | Consequent | *Support* | *Confidence* | *Lift* |
|---|---|---|---|---|---|
| 30 | P13 | P1 | 0.1309 | 0.6604 | 0.9534 |
| 30 | P16 | P1 | 0.0876 | 0.7491 | 1.0815 |
| 29 | P4 | P1 | 0.1772 | 0.6076 | 0.8772 |
| 16 | P66 | P71 | 0.0310 | 0.9146 | 26.6787 |
| 14 | P32 | P1 | 0.0037 | 1.0000 | 1.4436 |
| 14 | P73 | P1 | 0.0058 | 0.9333 | 1.3474 |
| 10 | P9,P16 | P1 | 0.0025 | 1.0000 | 1.4436 |
| 7 | P71 | P66 | 0.0310 | 0.9036 | 26.6787 |
| 6 | P6 | P18 | 0.0058 | 0.7000 | 39.4000 |
| 6 | P4,P16,P17 | P2 | 0.0008 | 1.0000 | 27.2000 |

Table 13 MOEAD-ARM-V2-PQR Bank

| Frequency | Antecedent | Consequent | *Confidence* | *Lift* | *Interestingness* |
|---|---|---|---|---|---|
| 8 | P71 | P66 | 0.9036 | 26.6787 | 0.8265 |
| 6 | P66 | P71 | 0.9146 | 26.6787 | 0.8265 |
| 4 | P21 | P20 | 0.7778 | 29.8889 | 0.2593 |
| 4 | P6 | P18 | 0.7000 | 39.4116 | 0.2279 |
| 3 | P20 | P21 | 0.3333 | 29.8889 | 0.2593 |
| 2 | P13 | P1 | 0.6604 | 0.9534 | 0.1248 |
| 2 | P73 | P1 | 0.9333 | 1.3474 | 0.0078 |
| 2 | P56 | P31 | 0.5000 | 41.7414 | 0.1207 |
| 2 | P24 | P54 | 0.2500 | 75.7000 | 0.0312 |
| 2 | P41 | P20 | 0.5556 | 21.3492 | 0.1323 |

Table 14 NSGA-III-ARM-V2-PQR Bank

| Frequency | Antecedent | Consequent | *Confidence* | *Lift* | *Interestingness* |
|---|---|---|---|---|---|
| 15 | P66 | P71 | 0.9146 | 26.6787 | 0.8265 |
| 8 | P71 | P66 | 0.9036 | 26.6787 | 0.8265 |
| 7 | P60 | P11 | 0.9286 | 9.1385 | 0.0491 |
| 6 | P4,P16,P17 | P2 | 1.0000 | 27.2000 | 0.0225 |
| 6 | P6 | P18 | 0.7000 | 39.4116 | 0.2279 |
| 6 | P21 | P20 | 0.7778 | 29.8889 | 0.2593 |
| 5 | P47 | P20 | 0.7083 | 27.2202 | 0.1911 |
| 5 | P1,P4,P32 | P2 | 1.0000 | 27.2000 | 0.0112 |
| 5 | P20 | P21 | 0.3333 | 29.8889 | 0.2593 |
| 4 | P1,P4,P49 | P2 | 1.0000 | 27.2000 | 0.0112 |

Table 15 MOEAD-ARM-V1-Grocery

| Frequency | Antecedent | Consequent | *Support* | *Confidence* | *Lift* |
|---|---|---|---|---|---|
| **3** | **apples,corned_b,hering,olives** | steak | 0.0969 | 0.8899 | 3.9242 |
| **2** | **avocado,cracker,ham,heineken** | artichok | 0.0989 | 0.9802 | 3.2170 |
| **2** | **apples,artichok,corned_b,hering,olives** | steak | 0.0140 | 1.0000 | 4.4097 |
| **2** | **avocado,corned_b,cracker,ham,heineken** | artichok | 0.0190 | 1.0000 | 3.2820 |
| **2** | **avocado,baguette,peppers,sardines** | apples | 0.0899 | 0.9890 | 3.1529 |
| **2** | **bourbon,chicken,corned_b,cracker** | peppers | 0.0919 | 0.9892 | 3.3454 |
| **2** | **apples,baguette,peppers,sardines** | avocado | 0.0899 | 1.0000 | 2.7576 |
| **2** | **apples,bourbon,chicken,corned_b,cracker** | peppers | 0.0200 | 1.0000 | 3.3818 |
| **1** | **apples,artichok,heineken,hering** | baguette | 0.0230 | 0.9583 | 2.4472 |
| **1** | **bourbon,coke,ice_crea,olives** | turkey | 0.0939 | 0.9691 | 3.4277 |

Table 16 NSGA-III-ARM-V1-Grocery

| Frequency | Antecedent | Consequent | *Support* | *Confidence* | *Lift* |
|---|---|---|---|---|---|
| 30 | soda | heineken | 0.2567 | 0.8082 | 1.3483 |
| 30 | cracker | heineken | 0.3656 | 0.7500 | 1.2513 |
| **30** | **baguette,cracker,soda** | heineken | 0.1259 | 0.9921 | 1.6552 |
| **30** | **cracker,Heineken** | soda | 0.2338 | 0.6393 | 2.0125 |
| **30** | **cracker,hering,soda** | heineken | 0.1369 | 0.9716 | 1.6210 |
| 30 | cracker,soda | heineken | 0.2338 | 0.9323 | 1.5553 |
| **30** | **cracker,heineken,hering** | soda | 0.1369 | 0.8405 | 2.6457 |
| **30** | **heineken,hering,soda** | cracker | 0.1369 | 0.9195 | 1.8860 |
| **30** | **heineken,soda** | cracker | 0.2338 | 0.9105 | 1.8677 |
| 30 | soda | cracker | 0.2507 | 0.7893 | 1.6191 |

Table 17 MOEAD-ARM-V2-Grocery

| Frequency | Antecedent | Consequent | *Confidence* | *Lift* | *Interestingness* |
|---|---|---|---|---|---|
| 1 | apples,artichok,avocado,baguette,bourbon,peppers | sardines | 1.0000 | 3.3818 | 0.0068 |
| 1 | apples,artichok,chicken,corned_b,hering,olives | steak | 1.0000 | 4.4097 | 0.0132 |
| 1 | baguette | avocado | 0.5485 | 1.5124 | 0.3248 |
| 1 | baguette,bordeaux,cracker,hering | heineken | 1.0000 | 1.6683 | 0.0083 |
| 1 | baguette,corned_b,ham,hering,ice_crea,olives | turkey | 1.0000 | 3.5400 | 0.0035 |
| 1 | bordeaux,chicken,heineken,ice_crea,sardines | coke | 1.0000 | 3.3818 | 0.0338 |
| 1 | bordeaux,chicken,heineken,sardines,steak | coke | 1.0000 | 3.3800 | 0.0034 |
| 1 | bordeaux,chicken,heineken,steak | coke | 1.0000 | 3.3800 | 0.0034 |
| 1 | bordeaux,chicken,ice_crea,sardines | coke | 1.0000 | 3.3818 | 0.0338 |
| 1 | bourbon,chicken,corned_b,cracker,ice_crea,soda | peppers | 1.0000 | 3.3818 | 0.0068 |

Table 18 NSGA-III-ARM-V2

| Freque | Antecedent | Consequent | *Confidence* | *Lift* | *Interestingness* |
|--------|-----------|------------|-------------|--------|-------------------|
| 29 | chicken,coke,heineken,ice_crea | sardines | 1.0000 | 3.3818 | 0.3918 |
| 29 | chicken,heineken,ice_crea,sardines | coke | 1.0000 | 3.3818 | 0.3918 |
| 29 | chicken,ice_crea,sardines | coke | 1.0000 | 3.3818 | 0.3918 |
| 27 | coke | ice_crea | 0.7432 | 2.3770 | 0.5223 |
| 17 | apples,corned_b,hering,olives | steak | 0.8899 | 3.9242 | 0.3802 |
| 17 | bourbon,coke,ice_crea,olives | turkey | 0.9691 | 3.4277 | 0.3219 |
| 16 | apples,artichok,corned_b,hering,olives | steak | 1.0000 | 4.4097 | 0.0617 |
| 16 | heineken,soda | cracker | 0.9105 | 1.8677 | 0.4365 |
| 15 | avocado,heineken | artichok | 0.7992 | 2.6229 | 0.5213 |
| 14 | apples,hering,olives,sardines | steak | 1.0000 | 4.4097 | 0.0529 |

Table 19 MOEAD-ARM-V1-Food

| Frequency | Antecedent | Consequent | *Support* | *Confidence* | *Lift* |
|-----------|-----------|------------|-----------|--------------|--------|
| 5 | fruitveg,freshmeat,cannedveg,cannedmeat,frozenmeal,beer,wine | dairy | 0.0010 | 1.0000 | 5.6500 |
| 5 | dairy,cannedveg,cannedmeat,frozenmeal,beer,wine | freshmeat | 0.0020 | 1.0000 | 5.4600 |
| 3 | freshmeat,cannedveg,cannedmeat,frozenmeal,beer,wine,softdrink | dairy | 0.0010 | 1.0000 | 5.6500 |
| 2 | dairy,cannedveg,frozenmeal,fish,confectionery | freshmeat | 0.0020 | 1.0000 | 5.4600 |
| 2 | freshmeat,cannedveg,cannedmeat,wine,fish,confectionery | softdrink | 0.0010 | 1.0000 | 5.4300 |
| 2 | freshmeat,cannedmeat,frozenmeal,wine,softdrink | dairy | 0.0010 | 1.0000 | 5.6500 |
| 1 | fruitveg,freshmeat,dairy,cannedveg,frozenmeal,beer | cannedmeat | 0.0010 | 1.0000 | 4.9000 |
| 1 | cannedveg,frozenmeal,softdrink,fish,confectionery | freshmeat | 0.0010 | 1.0000 | 5.4600 |
| 1 | dairy,cannedveg,cannedmeat,beer,wine | freshmeat | 0.0020 | 1.0000 | 5.4600 |
| 1 | freshmeat,cannedmeat,frozenmeal,wine,confectionery | beer | 0.0010 | 1.0000 | 3.4100 |

Table 20 NSGA-III-ARM-V1-Food

| Frequency | Antecedent | Consequent | Support | Confidence | Lift |
|---|---|---|---|---|---|
| **30** | **fruitveg,freshmeat,cannedveg,cannedmeat, frozenmeal,beer,wine** | dairy | 0.0010 | 1.0000 | 5.6500 |
| **30** | **fruitveg,cannedveg,frozenmeal,wine,fish** | beer | 0.0080 | 1.0000 | 3.4130 |
| 30 | frozenmeal | cannedveg | 0.1730 | 0.5728 | 1.8906 |
| 30 | cannedveg,beer | frozenmeal | 0.1460 | 0.8743 | 2.8949 |
| 30 | cannedveg,beer,fish | frozenmeal | 0.0440 | 0.9167 | 3.0353 |
| **30** | **cannedveg,cannedmeat,beer,fish** | frozenmeal | 0.0120 | 1.0000 | 3.3113 |
| **30** | **freshmeat,frozenmeal,beer** | cannedveg | 0.0290 | 0.9667 | 3.1903 |
| **30** | **freshmeat,cannedveg,frozenmeal,wine, softdrink** | dairy | 0.0010 | 1.0000 | 5.6500 |
| **30** | **fruitveg,freshmeat,cannedveg,cannedmeat, frozenmeal,wine** | dairy | 0.0010 | 1.0000 | 5.6500 |
| **30** | **freshmeat,cannedveg,cannedmeat,frozenmeal ,wine,softdrink** | dairy | 0.0010 | 1.0000 | 5.6500 |

Table 21 MOEAD-ARM-V2-Food

| Freq | Antecedent | Consequent | Confidence | Lift | Interestingness |
|---|---|---|---|---|---|
| 5 | Napoleon,Lemon_T3,Blueberry_T2,Raspberry_T2 | Pecan | 0.0010 | 1.0000 | 25.0000 |
| 5 | fruitveg,freshmeat,cannedveg,cannedmeat,frozenmeal, beer,wine | dairy | 1.0000 | 5.6500 | 0.0057 |
| 3 | freshmeat,cannedveg,cannedmeat,frozenmeal,beer,win e,softdrink | dairy | 1.0000 | 5.6500 | 0.0057 |
| 2 | freshmeat,cannedveg,cannedmeat,frozenmeal,wine,soft drink | dairy | 1.0000 | 5.6500 | 0.0057 |
| 2 | fruitveg,freshmeat,beer,wine,fish,confectionery | dairy | 1.0000 | 5.6500 | 0.0057 |
| 2 | fruitveg,cannedveg,frozenmeal,beer,wine,fish,confectio nery | cannedm eat | 1.0000 | 4.9000 | 0.0049 |
| 2 | freshmeat,dairy,cannedveg,frozenmeal,beer,softdrink | cannedm eat | 1.0000 | 4.9000 | 0.0049 |
| 2 | dairy,cannedveg,cannedmeat,frozenmeal,beer,wine | freshmea t | 1.0000 | 5.4645 | 0.0109 |
| 2 | freshmeat,cannedveg,cannedmeat,beer,wine,softdrink | dairy | 1.0000 | 5.6500 | 0.0057 |
| 1 | freshmeat,cannedmeat,beer,wine,softdrink | dairy | 1.0000 | 5.6500 | 0.0057 |
| 1 | beer | frozenme al | 0.0010 | 1.0000 | 25.0000 |

Table 22 NSGA-III-ARM-V2-Food

| Freq | Antecedent | Consequent | Confidence | Lift | Interestingness |
|---|---|---|---|---|---|
| 30 | fruitveg,freshmeat,cannedveg,cannedmeat,frozenmeal,beer,wine | dairy | 1.0000 | 5.6500 | 0.0057 |
| 30 | fruitveg,cannedveg,frozenmeal,wine,fish | beer | 1.0000 | 3.4130 | 0.0273 |
| 30 | cannedveg,beer,fish | frozenmeal | 0.9167 | 3.0353 | 0.1335 |
| 30 | cannedveg,cannedmeat,beer,fish | frozenmeal | 1.0000 | 3.3113 | 0.0397 |
| 30 | freshmeat,frozenmeal,beer | cannedveg | 0.9667 | 3.1903 | 0.0925 |
| 30 | freshmeat,cannedveg,frozenmeal,beer,wine,softdrink | dairy | 1.0000 | 5.6500 | 0.0057 |
| 30 | freshmeat,cannedveg,cannedmeat,frozenmeal,beer,wine,softdrink | dairy | 1.0000 | 5.6500 | 0.0057 |
| 30 | fruitveg,cannedveg,frozenmeal,wine | beer | 1.0000 | 3.4130 | 0.0375 |
| 30 | cannedveg,beer | frozenmeal | 0.8743 | 2.8949 | 0.4226 |
| 30 | fruitveg,freshmeat,cannedveg,frozenmeal,beer,wine | dairy | 1.0000 | 5.6500 | 0.0057 |

Table 23 MOEAD-ARM-V1-Books

| Freq | Antecedent | Consequent | Support | Confidence | Lift |
|---|---|---|---|---|---|
| 4 | CookBks,DoItYBks,RefBks,ArtBks,GeogBks,ItalCook,ItalArt,Florence | ItalAtlas | 0.0010 | 1.0000 | 27.0000 |
| 3 | ChildBks,YouthBks,CookBks,DoItYBks,RefBks,ArtBks,GeogBks,ItalCook,ItalArt,Florence | ItalAtlas | 0.0010 | 1.0000 | 27.0000 |
| 3 | YouthBks,CookBks,DoItYBks,RefBks,ArtBks,GeogBks,ItalCook,ItalArt,Florence | ItalAtlas | 0.0010 | 1.0000 | 27.0000 |
| 3 | YouthBks,DoItYBks,RefBks,GeogBks,ItalCook,ItalArt,Florence | ItalAtlas | 0.0010 | 1.0000 | 27.0000 |
| 2 | YouthBks,DoItYBks,RefBks,ItalCook,ItalArt,Florence | ItalAtlas | 0.0010 | 1.0000 | 27.0000 |
| 2 | YouthBks,DoItYBks,RefBks,ArtBks,GeogBks,ItalCook,ItalArt,Florence | ItalAtlas | 0.0010 | 1.0000 | 27.0000 |
| 2 | ChildBks,YouthBks,CookBks,DoItYBks,RefBks,ArtBks,ItalCook,ItalArt,Florence | ItalAtlas | 0.0010 | 1.0000 | 27.0000 |
| 2 | DoItYBks,RefBks,GeogBks,ItalArt,Florence | ItalAtlas | 0.0010 | 1.0000 | 27.0000 |
| 2 | ChildBks,YouthBks,CookBks,DoItYBks,RefBks,GeogBks,ItalCook,ItalArt,Florence | ItalAtlas | 0.0010 | 1.0000 | 27.0000 |
| 1 | ChildBks,YouthBks,ArtBks,GeogBks,ItalAtlas | ItalArt | 0.0045 | 1.0000 | 20.6000 |

Table 24 NSGA-III-ARM-V1-Books

| Frequenc | Antecedent | Consequent | Support | Confidence | Lift |
|---|---|---|---|---|---|
| 30 | ChildBks,YouthBks,CookBks | RefBks | 0.0680 | 0.5271 | 2.4575 |
| 30 | ChildBks,CookBks | ItalCook | 0.0850 | 0.3320 | 2.9254 |
| 30 | ArtBks,GeogBks,ItalAtlas | ItalArt | 0.0115 | 1.0000 | 20.6000 |
| 30 | ItalAtlas | RefBks | 0.0370 | 1.0000 | 4.6620 |
| 30 | RefBks,ItalArt | ItalAtlas | 0.0165 | 0.8250 | 22.3000 |
| 30 | RefBks,ArtBks,ItalArt | ItalAtlas | 0.0165 | 0.8250 | 22.3000 |
| 30 | RefBks,ArtBks,ItalAtlas | ItalArt | 0.0165 | 0.9170 | 18.9000 |
| 30 | RefBks,ArtBks,GeogBks,ItalAtlas | ItalArt | 0.0115 | 1.0000 | 20.6000 |
| 30 | DoItYBks,RefBks,GeogBks,ItalArt | ItalAtlas | 0.0055 | 0.9170 | 24.8000 |
| 30 | DoItYBks,RefBks,ArtBks,GeogBks,ItalArt | ItalAtlas | 0.0055 | 0.9170 | 24.8000 |

Table 25 MOEAD-ARM-V2-Books

| Frequency | Antecedent | Consequent | Confidence | Lift | Interestingness |
|---|---|---|---|---|---|
| 5 | ItalArt | ItalCook | 0.7732 | 6.8123 | 0.2555 |
| 3 | ItalArt | ItalAtlas | 0.3402 | 9.1948 | 0.1517 |
| 3 | ItalAtlas | ItalArt | 0.4459 | 9.1948 | 0.1517 |
| 3 | DoItYBks | CookBks | 0.6649 | 1.5427 | 0.2892 |
| 2 | ItalAtlas | RefBks | 1.0000 | 4.6620 | 0.1725 |
| 2 | CookBks | RefBks | 0.3538 | 1.6495 | 0.2515 |
| 2 | YouthBks,ItalArt | ItalAtlas | 0.3696 | 9.9882 | 0.0849 |
| 2 | CookBks | GeogBks | 0.4466 | 1.6182 | 0.3115 |
| 2 | ArtBks | DoItYBks | 0.5124 | 1.8172 | 0.2244 |
| 2 | DoItYBks,ItalCook | ItalArt | 0.4274 | 8.8113 | 0.2203 |

Table 26 NSGA-III-ARM-V2-Books

| Freque ncy | Antecedent | Consequ ent | Confid ence | Lift | Interesting ness |
|---|---|---|---|---|---|
| 30 | ChildBks,YouthBks,DoItYBks,RefBks,Art Bks,ItalArt,Florence | ItalAtlas | 1.0000 | 27.0270 | 0.0405 |
| 30 | DoItYBks,RefBks,ItalArt | ItalAtlas | 0.8636 | 23.3415 | 0.2217 |
| 30 | ArtBks,ItalCook | ItalArt | 0.6637 | 13.6849 | 0.5132 |
| 30 | ArtBks,ItalCook,ItalAtlas | ItalArt | 0.9615 | 19.8255 | 0.2478 |
| 30 | ArtBks,GeogBks,ItalAtlas | ItalArt | 1.0000 | 20.6186 | 0.2371 |
| 30 | RefBks,ItalArt | ItalAtlas | 0.8250 | 22.2973 | 0.3679 |
| 30 | RefBks,GeogBks,ItalArt | ItalAtlas | 0.8519 | 23.0230 | 0.2648 |
| 30 | RefBks,ArtBks,ItalArt | ItalAtlas | 0.8250 | 22.2973 | 0.3679 |
| 30 | RefBks,ArtBks,ItalAtlas | ItalArt | 0.9167 | 18.9003 | 0.3119 |
| 30 | RefBks,ArtBks,ItalCook,ItalAtlas | ItalArt | 0.9615 | 19.8255 | 0.2478 |

Table 27 MOEAD-ARM-V1-XYZbank

| Frequency | Antecedent | Consequent | Support | Confidence | Lift |
|---|---|---|---|---|---|
| 15 | AGL1 | SB2 | 0.0187 | 0.7215 | 3.1729 |
| 13 | FD1 | SB1 | 0.0660 | 0.4672 | 1.0307 |
| **11** | **GL3,SB1** | GL1 | 0.0158 | 0.4812 | 6.2938 |
| **8** | **P8,P9** | P10 | 0.0107 | 0.8680 | 24.8000 |
| **8** | **GL3** | SB2 | 0.0231 | 0.2518 | 1.1072 |
| **7** | **SB1** | FD1 | 0.0660 | 0.1455 | 1.0307 |
| **5** | **GL3** | P4 | 0.0233 | 0.2536 | 1.8097 |
| **5** | **P4** | GL3 | 0.0233 | 0.1663 | 1.8097 |
| **4** | **GL1,GL3,SB1** | P1 | 0.0064 | 0.4060 | 7.3700 |
| **4** | **GL1** | P4 | 0.0221 | 0.2886 | 2.0599 |

Table 28 NSGA-III-ARM-V1-XYZbank

| Frequency | Antecedent | Consequent | *Support* | *Confidence* | *Lift* |
|---|---|---|---|---|---|
| 22 | FD1 | SB1 | 0.0660 | 0.4672 | 1.0307 |
| 21 | AGL1 | SB2 | 0.0187 | 0.7215 | 3.1729 |
| 11 | **P8,P9** | P10 | 0.0107 | 0.8680 | 24.8000 |
| **10** | **GL3,SB1** | GL1 | 0.0158 | 0.4812 | 6.2938 |
| **9** | **GL3** | SB2 | 0.0231 | 0.2518 | 1.1072 |
| **8** | **GL3** | P4 | 0.0233 | 0.2536 | 1.8097 |
| **7** | **P4** | GL1 | 0.0221 | 0.1575 | 2.0599 |
| **6** | **SB1** | FD1 | 0.0660 | 0.1455 | 1.0307 |
| **6** | **GL1** | P4 | 0.0221 | 0.2886 | 2.0599 |
| **6** | **P4,GL3** | GL1 | 0.0127 | 0.5458 | 7.1384 |

Table 29 MOEAD-ARM-V2-XYZbank

| Frequency | Antecedent | Consequent | *Confidence* | *Lift* | *Interestingness* |
|---|---|---|---|---|---|
| 9 | P8,P9 | P10 | 0.8675 | 24.8250 | 0.2668 |
| 6 | AGL1 | SB2 | 0.7215 | 3.1729 | 0.0593 |
| 3 | GL3,SB1 | GL1 | 0.4812 | 6.2938 | 0.0991 |
| 3 | P8,P10 | P9 | 0.8506 | 17.3402 | 0.1863 |
| 3 | SB2,P1 | GL3 | 0.6044 | 6.5782 | 0.0594 |
| 3 | P4,GL3 | P1 | 0.3908 | 7.0899 | 0.0646 |
| 3 | GL1,P4,P1 | GL3 | 0.8111 | 8.8281 | 0.0529 |
| 2 | P4,P9 | P10 | 0.6185 | 17.6983 | 0.1554 |
| 2 | P2,FD1,P6 | P3 | 0.1670 | 2030.0000 | 0.1670 |
| 2 | FD1 | SB1 | 0.4672 | 1.0307 | 0.0680 |

Table 30 NSGA-III-ARM-V2-XYZbank

| Frequency | Antecedent | Consequent | *Confidence* | *Lift* | *Interestingness* |
|---|---|---|---|---|---|
| 9 | P8,P9 | P10 | 0.8675 | 24.8250 | 0.2668 |
| 5 | AGL1 | SB2 | 0.7215 | 3.1729 | 0.0593 |
| 4 | P8,P10 | P9 | 0.8506 | 17.3402 | 0.1863 |
| 4 | GL3,SB1 | GL1 | 0.4812 | 6.2938 | 0.0991 |
| 3 | SB2,P1 | GL3 | 0.6044 | 6.5782 | 0.0594 |
| 3 | GL1,P4,P1 | GL3 | 0.8111 | 8.8281 | 0.0529 |
| 3 | P4,GL3 | P1 | 0.3908 | 7.0899 | 0.0646 |
| 2 | P5,P7,SB1 | P2 | 0.6670 | 26.8000 | 0.0044 |
| 2 | P2,FD1,P6 | P3 | 0.1670 | 2030.0000 | 0.1670 |
| 2 | P4,P10,FD1 | P9 | 0.7917 | 16.1378 | 0.0503 |